# Leveraging Generative AI Through Prompt Engineering and Rigorous Validation to Create Comprehensive Synthetic Datasets for AI Training in Healthcare

Polycarp Nalela[1*]

[1*]Institute for Data Science and Informatics, University of Missouri, 22 Heinkel Building, Columbia, 65201, MO, USA.

Corresponding author(s). E-mail(s): polycarpnalela@missouri.edu;

**Abstract**

Access to high-quality medical data is often restricted due to privacy concerns, posing significant challenges for training artificial intelligence (AI) algorithms within Electronic Health Record (EHR) applications. In this study, prompt engineering with the GPT-4 API was employed to generate high-quality synthetic datasets aimed at overcoming this limitation. The generated data encompassed a comprehensive array of patient admission information, including healthcare provider details, hospital departments, wards, bed assignments, patient demographics, emergency contacts, vital signs, immunizations, allergies, medical histories, appointments, hospital visits, laboratory tests, diagnoses, treatment plans, medications, clinical notes, visit logs, discharge summaries, and referrals. To ensure data quality and integrity, advanced validation techniques were implemented utilizing models such as BERT's Next Sentence Prediction for sentence coherence, GPT-2 for overall plausibility, RoBERTa for logical consistency, autoencoders for anomaly detection, and conducted diversity analysis. Synthetic data that met all validation criteria were integrated into a comprehensive PostgreSQL database, serving as the data management system for the EHR application. This approach demonstrates that leveraging generative AI models with rigorous validation can effectively produce high-quality synthetic medical data, facilitating the training of AI algorithms while addressing privacy concerns associated with real patient data.

# 1 Introduction

## 1.1 Background

Artificial intelligence (AI) has emerged as a transformative force in healthcare, offering the potential to enhance diagnostics, personalize treatment plans, and improve patient outcomes [1]. Central to the success of AI algorithms, particularly in machine learning and deep learning applications, is the availability of large, high-quality datasets for training and validation [2]. However, access to such datasets in the healthcare domain is often constrained by stringent privacy regulations and ethical considerations, such as the Health Insurance Portability and Accountability Act (HIPAA) in the United States and the General Data Protection Regulation (GDPR) in the European Union [3, 4]. These regulations limit the sharing and utilization of patient data, posing significant challenges for researchers and developers aiming to leverage AI technologies in electronic health record (EHR) applications.

To address these challenges, synthetic data generation has gained attention as a viable alternative to real patient data [5]. Synthetic datasets mimic the statistical properties of real-world data without compromising patient privacy, enabling the development and testing of AI algorithms [6]. Traditional methods for generating synthetic medical data include statistical modeling and data augmentation techniques [7]. However, these approaches often fail to capture the complex, high-dimensional relationships inherent in healthcare data, potentially leading to less effective AI models [8].

Recent advancements in generative AI models, such as Generative Adversarial Networks (GANs) and transformer-based architectures like GPT-3 and GPT-4, have shown promise in producing high-fidelity synthetic data [9, 10]. These models can generate data that closely resembles real patient information, encompassing various aspects such as demographics, clinical notes, and laboratory results. Nevertheless, generating high-quality synthetic datasets using these models requires careful prompt engineering and rigorous validation to ensure data quality and integrity [11].

## 1.2 Prompt Engineering

Prompt engineering involves crafting input queries or prompts that guide the generative AI models to produce desired outputs [11]. In the context of synthetic data generation for healthcare, prompt engineering is crucial for specifying the structure, content, and diversity of the generated data. Moreover, rigorous validation techniques are essential to assess the plausibility, coherence, and statistical fidelity of the synthetic data [12]. This evaluation check is more relevant when populating an entire healthcare database having interconnected tables as the slightest inconsistency in the data can results in adverse decisions from AI models trained on such data sets. Advanced validation models, such as BERT's Next Sentence Prediction for sentence coherence [13], GPT-2 for overall plausibility [14], RoBERTa for logical consistency [15], and autoencoders for anomaly detection [16], can be employed to ensure the integrity of the generated datasets.

## 1.3 Coherence, Plausibility, and Consistency Checks

Ensuring the quality of synthetic healthcare data necessitates a multifaceted validation approach that includes coherence, plausibility, and consistency checks. **Coherence** refers to the logical flow and meaningful connections within the data. For example, in a patient's clinical narrative, the progression from symptoms to diagnosis and then to treatment should follow a medically logical sequence. A lack of coherence might manifest as a patient being treated for a condition that was not previously diagnosed or mentioned in their medical history.

**Plausibility** involves verifying that the generated data is medically reasonable and aligns with known clinical knowledge. For instance, a synthetic patient record indicating a 30-year-old male diagnosed with osteoporosis would prompt scrutiny, as osteoporosis is more prevalent in older adults and females [17]. Ensuring plausibility helps prevent the inclusion of unrealistic or impossible scenarios that could mislead AI models during training.

**Consistency checks** focus on identifying and correcting contradictions within the data. An example is a patient record where the allergy section lists a penicillin allergy, yet the medication list includes a penicillin-based antibiotic. Such inconsistencies can confuse AI algorithms, leading to inaccurate predictions or recommendations [18]. Thus, enforcing consistency maintains the reliability of the synthetic dataset.

## 1.4 Anomaly Detection

Anomaly detection is critical for identifying data points that deviate significantly from expected patterns, which may indicate errors in the data generation process. In the medical context, anomalies might include vital sign readings that are physiologically impossible, such as a diastolic blood pressure of zero, or lab results that are orders of magnitude outside normal ranges. For example, a recorded potassium level of 15 mmol/L is abnormally high and likely indicative of a data generation error, as such a level is incompatible with life [19]. Detecting and addressing these anomalies ensures that the synthetic data does not contain misleading or harmful information.

## 1.5 Diversity Analysis

Diversity analysis ensures that the synthetic dataset captures a wide range of patient demographics, medical conditions, and clinical scenarios, mirroring the variability found in real-world populations. This is crucial for training AI models that are robust and generalizable across different patient groups. For instance, including a diverse age range in the synthetic dataset allows AI algorithms to learn patterns relevant to pediatric, adult, and geriatric populations [20]. Similarly, representing various ethnic backgrounds can help the AI model account for genetic predispositions to certain diseases [21]. Performing diversity analysis aims to prevent biases in AI models that could arise from homogeneous training data.

In this study, the use of prompt engineering with the GPT-4 API is explore to generate comprehensive synthetic datasets for AI training in healthcare. The approach focuses on creating a wide range of patient admission information, including healthcare provider details, hospital departments, patient demographics, vital signs, medical histories, laboratory tests, and more. By implementing advanced validation techniques and conducting diversity analysis, the study demonstrates that it is possible to pro- duce high-quality synthetic data that can facilitate the training of AI algorithms while addressing privacy concerns associated with real patient data.

The contributions from this study are as follows:

- **Development of a Prompt Engineering Framework:** The study develops a design for a systematic approach for crafting prompts that guide the GPT-4 model in generating comprehensive and realistic synthetic healthcare data.
- **Implementation of Rigorous Validation Techniques:** The study utilizes state-of-the-art models and statistical methods to validate the quality, coherence, and diversity of the synthetic data.
- **Integration into a Data Management System:** The study incorporate the validated synthetic data into a PostgreSQL database, establishing a robust data management system suitable for EHR applications.

## 2 Methodology

### 2.1 Database Design

The implementation the Electronic Health Record (EHR) system began by designing the database for storage and management of the synthetic data. PostgreSQL was employed as the Database Management System (DBMS) due to its robustness, scalability, and adherence to SQL standards. PostgreSQL's advanced capabilities in handling complex queries and supporting extensive data types made it well suited for managing the intricate and voluminous data inherent in healthcare records. The database was meticulously designed to encompass a comprehensive range of information, structured into 22 interrelated tables that captured various facets of patient care and hospital operations.

These tables included data on hospital staff, departments, wards, beds, patient details, emergency contacts, vital signs, immunizations, allergies, medical histories, appointments, hospital visits, test results, diagnoses, admissions, treatment plans, medications, clinical notes, visit logs, discharge summaries, referrals, and billing information. The relational structure of the database ensured data integrity and facilitated efficient data retrieval and analysis, which was essential for the subsequent training of AI algorithms. This way, the synthetic data generated through prompt engineering with the GPT-4 API would seamlessly be integrated into the PostgreSQL database, providing a realistic and privacy-compliant dataset for developing and testing AI applications within the EHR system.

### 2.2 Synthetic Data Generation

Prompt engineering techniques with the GPT-4 API were utilized to populate the database with realistic synthetic data. Prompt engineering involves crafting specific inputs to guide the AI model in generating desired outputs with high accuracy and relevance [11]. Customized prompts were developed for each of the 22 database tables to ensure that the generated data reflected real-world healthcare information accurately.

For example, to generate patient details, prompts included parameters for demographic diversity, such as varying ages, genders, ethnicities, and geographic locations. Clinical notes required complex prompts that instructed the model to create coherent medical narratives, including symptoms, diagnoses, treatments, and follow-up plans. Iterative refining these prompts and reviewing the outputs, ensured that the synthetic data was both diverse and representative of actual patient records.

## 2.3 Data Validation and Quality Assurance

Ensuring the quality and reliability of the synthetic data was crucial for the effectiveness of the AI models intended for training on this dataset. After integrating data from 42 patients across the 22 tables, we obtained a robust dataset comprising 445,500 records and 49 selected features. This extensive dataset provided a solid foundation for thorough validation and evaluation, reflecting a wide range of clinical scenarios and patient demographics. Each step was designed to scrutinize the data from different angles, employing advanced models and statistical techniques to ensure the synthetic data closely resembled real-world healthcare records.

The **coherence assessment** aimed to evaluate the logical flow and structure of textual data within each record. For instance, in a patient's clinical notes, it was crucial that the description of symptoms logically led to the diagnosis and subsequent treatment plan. Coherence assessment was achieved using BERT's Next Sentence Prediction (NSP) model, which evaluates whether two consecutive pieces of text logically follow one another. For each record, sentences representing vital information, such as body weight, height, and blood pressure, were extracted and paired with sentences describing chronic conditions, past surgeries, family medical history, diagnosis, treat- ment, admission, etc. These pairs were tokenized and processed through BERT NSP, leveraging the computational efficiency of GPUs when available. The model returned a probability distribution over two labels: "IsNext" and "NotNext," corresponding to coherent and incoherent relationships, respectively. By evaluating these probabilities, records with incoherent sentence sequences were flagged for review. The accuracy of coherence checks was further visualized by plotting sentence transition probabilities, which provided an overview of the logical flow across records, ensuring that medical narratives maintained a consistent structure.

In the **plausibility assessment**, the aim was to determine if the content was realistic and believable based on general and domain-specific medical knowledge. GPT-2 was utilized for evaluating the natural language consistency and realism of medical records. For each record, textual information, including admission reasons, diagnoses, conditions, and prescribed medications, was combined into a coherent narrative. This text was tokenized and processed through GPT-2 to calculate perplexity scores, a measure of how well the language aligns with natural patterns found in medical documentation. Lower perplexity scores indicated higher plausibility. The distribution of perplexity scores across records was visualized using histograms, with a predefined threshold (95th percentile) demarcating realistic records from implausible ones. Records with perplexity scores significantly exceeding this threshold were flagged as containing unrealistic or nonsensical content. By leveraging GPU acceleration, this method ensured efficient evaluation of large datasets, producing high-quality records aligned with known medical standards.

The **consistency checks** ensured that related fields within a medical record aligned logically according to domain-specific rules. This was achieved using RoBERTa's Natural Language Inference (NLI) model, which evaluates relationships such as entailment, contradiction, and neutrality between a premise and a hypothe- sis. For example, premises describing patient vitals, such as blood group, height, and weight, were compared with hypotheses about their severity classification or expected health outcomes; or if a patient's record indicated an allergy to penicillin (premise), the hypothesis would be that their medication list should not include penicillin- based antibiotics. The inputs were tokenized and processed through the NLI model, which outputted probabilities for each possible relationship. Records labeled as con- tradictions were flagged for inconsistency, while entailment or neutral labels were considered consistent. The consistency score distributions were visualized to highlight patterns across the dataset, and combined anomaly scores incorporating consistency and plausibility were analyzed. This approach ensured logical alignment across fields, contributing to the dataset's overall integrity and utility for AI training.

**Anomaly detection using autoencoders** was a critical component of ensuring the reliability of the synthetic dataset, leveraging advanced neural network tech- niques to identify outliers effectively. Initial step was the automatic distinguishing of numerical features, such as integer and float columns, and categorical features. These distinctions ensured precise handling of the data during preprocessing. Missing values were addressed by removing incomplete entries to maintain data integrity.

The preprocessing pipeline consisted of standardizing numerical features using StandardScaler and encoding categorical features with OneHotEncoder. This ensured that numerical attributes were normalized to have zero mean and unit variance, while categorical variables were transformed into a sparse, interpretable format. The trans- formed dataset was then represented in a unified structure, where numerical and encoded categorical features were combined, creating a high-dimensional input matrix.

The anomaly detection model was implemented using an autoencoder, a neural network architecture designed for unsupervised learning of data representations. Input features were compressed into a lower-dimensional latent space through encoding lay- ers with progressively fewer neurons, followed by reconstruction layers that expanded the representation back to the original dimensionality. The model was compiled with the Mean Squared Error (MSE) loss function to minimize reconstruction errors dur- ing training. By training the autoencoder on a GPU when available, computational efficiency was enhanced for high-dimensional healthcare data.

The reconstruction error for each data point was computed as the MSE between the original and reconstructed inputs. A threshold for anomalies was defined dynamically as the mean reconstruction error plus two standard deviations, capturing significant deviations from the dataset's normal patterns. Records exceeding this threshold were flagged as anomalies, and detailed statistics, including the reconstruction error for each instance, were appended to the dataset.

The identified anomalies were further examined by visualizing reconstruction error distributions, revealing patterns that diverged significantly from typical data behavior.

This rigorous process not only isolated erroneous records but also provided insights into potential gaps or inconsistencies in the data generation pipeline, enabling iterative refinement of the synthetic dataset. This comprehensive approach ensured that the final dataset was robust, coherent, and free from significant deviations that could undermine AI model training.

Lastly, the **diversity and coverage analysis** ensured that the synthetic data encompassed a wide range of possible values and combinations, reflecting real-world variability. Coverage metrics were calculated for categorical variables to assess how well different categories were represented. Diversity indices, such as the Shannon Diver‐sity Index, were computed for features like diagnoses, treatments, and demographic attributes. Higher diversity indices indicated a broader coverage, confirming that the synthetic data captured the heterogeneity necessary for training AI models that are generalizable and unbiased.

This comprehensive validation framework, ensured that the synthetic data was not only realistic and coherent but also diverse and consistent with medical knowledge and practices. The metrics and visualizations generated provided quantitative and qualitative insights into the data quality, guiding iterative refinements in the data generation process. This rigorous approach was essential for producing a high-quality synthetic dataset suitable for developing and testing AI applications in a privacy-compliant manner within the EHR system.

## 2.4 Integration into the EHR System

After validation, the synthetic data was seamlessly integrated into the PostgreSQL database. Data import scripts mapped the generated data to the corresponding tables and fields, adhering to the database schema's data types and constraints. Referential integrity was maintained by establishing proper relationships between tables using primary and foreign keys.

The integration of validated synthetic data into the EHR system facilitated efficient data retrieval and management. This comprehensive, realistic dataset provided a solid foundation for developing and testing AI applications within a secure and privacy-compliant environment.

# 3 Results

## 3.1 Coherence assessment

The analysis of the Next Sentence Prediction (NSP) probabilities revealed a highly coherent dataset. The least observed average probability for NSP was 0.99800, a value that indicates minimal incoherence in the logical flow of textual records. The majority of the NSP probabilities were concentrated at 1.00000, signifying that the sentences within the medical narratives and other textual components demonstrated a nearly perfect logical progression. This outcome reflects the effectiveness of BERT's Next Sentence Prediction model in evaluating coherence across patient records. The visual‐ization (Figure 1) further underscores this observation, as the frequency distribution

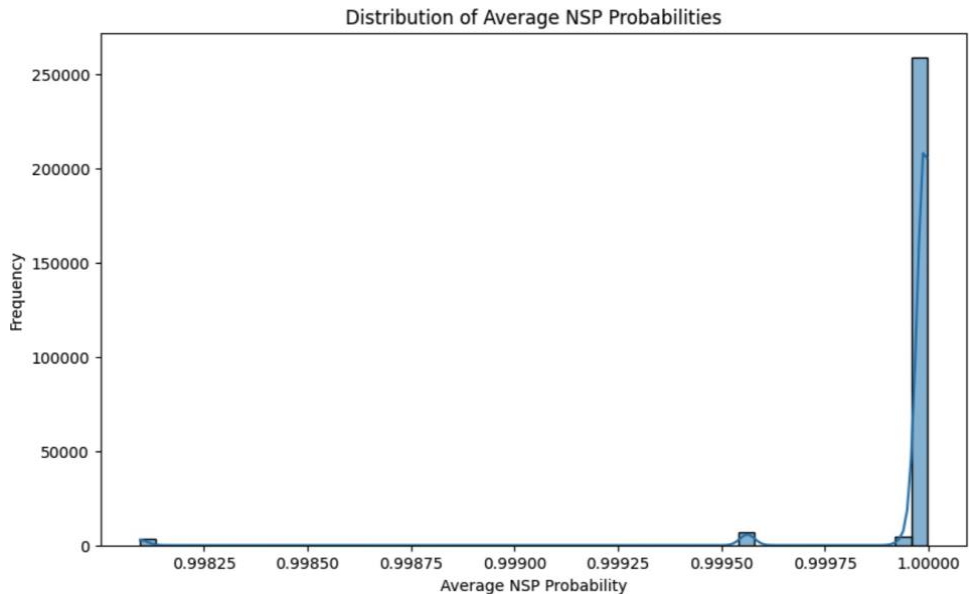
**Fig. 1**: Distribution of Average Next Sentence Probabilities

shows an overwhelming concentration at the highest probability range, with negligi- ble deviations. This result confirms that the generated records maintain a consistent and logical narrative structure, aligning with the expected coherence for real-world healthcare data.

## 3.2 Plausibility assessment

The distribution of perplexity scores offered further insights into the plausibility of the generated synthetic records. Using GPT-2, the perplexity scores were evaluated to measure how well the textual data aligned with natural language patterns observed in medical documentation. From this analysis, 22,950 records fell below the prede- fined perplexity threshold, as indicated by the red dashed line in the visualization (Figure 2). The scores ranged broadly between approximately 10 and 120, with a sig- nificant concentration observed between 20 and 50, suggesting that the majority of the generated records were realistic and plausible. The clustering of scores in this range highlights that the synthetic data closely adheres to natural language conven- tions. Only a small fraction of records surpassed the threshold, which were flagged for further scrutiny, while the bulk of the dataset exhibited high alignment with known medical text patterns.

## 3.3 Anomaly assessment with autoencoder

For the reconstruction error analysis, an autoencoder was employed to identify anoma- lies within the numerical and categorical fields of the dataset. The analysis revealed that a total of 40,490 records were flagged as anomalies due to reconstruction errors exceeding the dynamically calculated threshold. Despite these anomalies, the majority of the reconstruction errors were tightly clustered below 0.08, with a sharp decline in frequency as the errors approached the threshold. A smaller subset of records exhibited higher reconstruction errors, reaching up to 0.12, which were identified as outliers. The visualization (Figure 3) highlights this concentration of errors, with most records falling well within the acceptable range, reaffirming the robustness of the synthetic data generation pipeline in producing reliable and realistic numerical features.

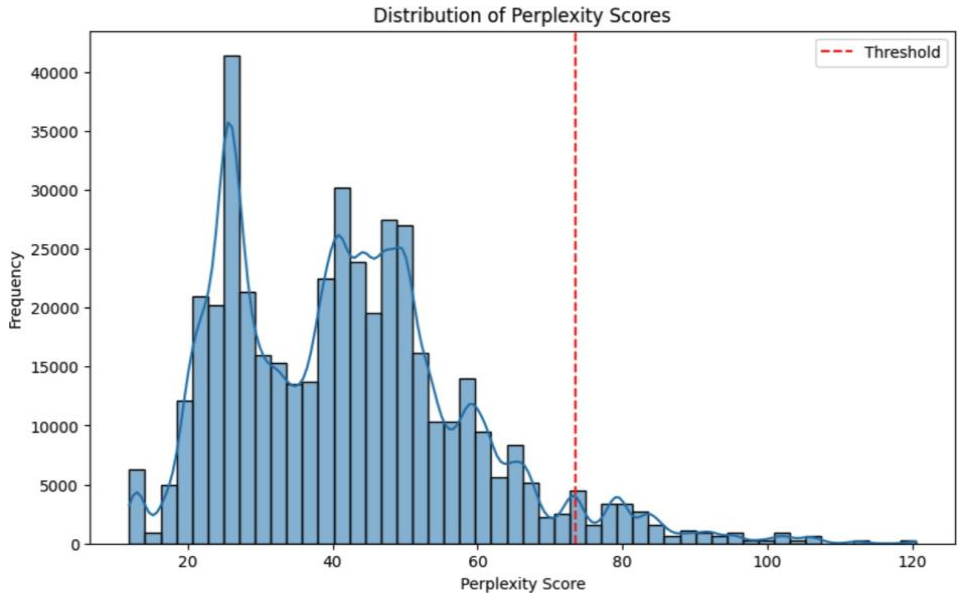

**Fig. 2**: Distribution of perplexity scores

### 3.4 Consistency assessment

The consistency scores, derived using RoBERTa's Natural Language Inference (NLI) model, provided a measure of the logical alignment between related fields within the records. The analysis revealed that the lowest observed consistency score (probability of entailment) was 0.9750, a value that suggests only minor instances of inconsistency. The majority of the records exhibited consistency scores between 0.9875 and 0.9925, as visualized in the distribution plot (Figure 4). This tight clustering at the higher end of the score range indicates that most relationships within the dataset, such as correlations between allergies and prescribed medications or symptoms and diagnoses, were logically consistent. The smooth distribution further validates the ability of the validation framework to identify and correct contradictions effectively, ensuring logical integrity across the dataset.

The combined anomaly scores provided a holistic view of the anomalies flagged through multiple validation checks. The scores ranged from 10 to 120, with a majority of the records clustering between 20 and 60, and peaks observed around 30 to 40. This distribution reflects localized deviations that were captured during the integration of coherence, plausibility, and anomaly detection results. Records with higher combined scores were identified as those requiring further attention, while the majority of the dataset maintained acceptable anomaly levels. Figure 5 illustrates these findings with a smooth frequency distribution, highlighting the effectiveness of combining multiple validation metrics to ensure data quality.

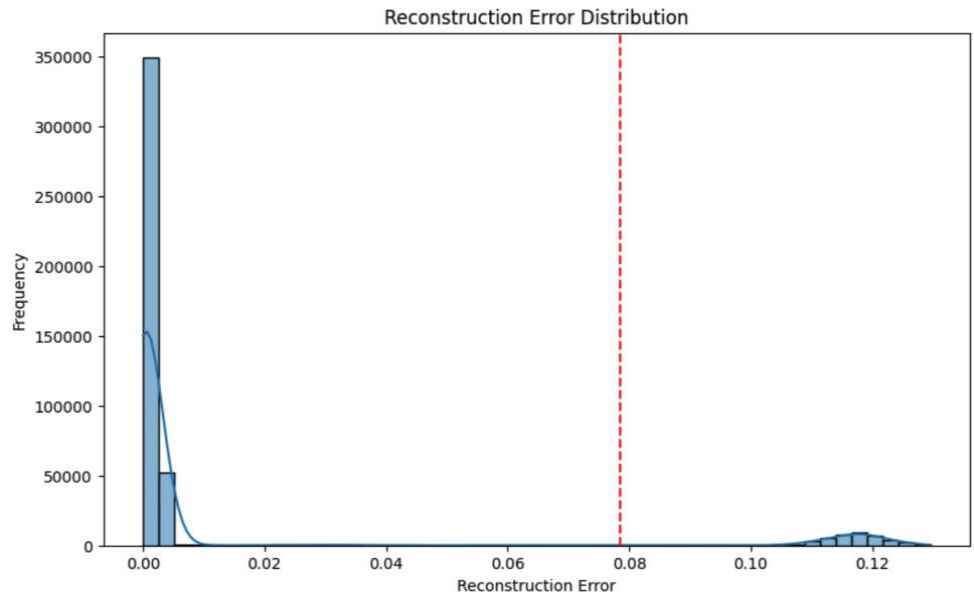

**Fig. 3**: Distribution of reconstruction error

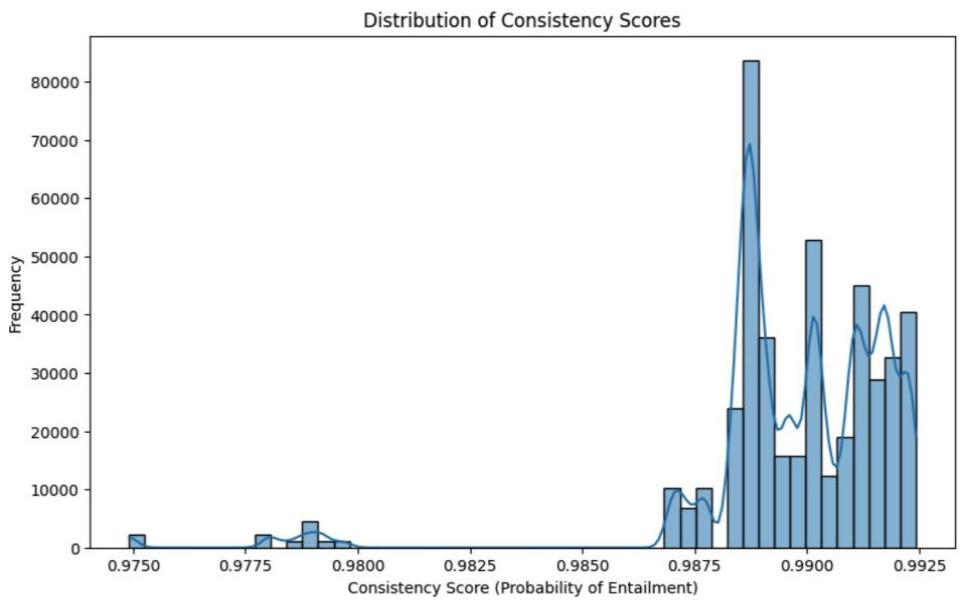

**Fig. 4**: Distribution of consistency scores

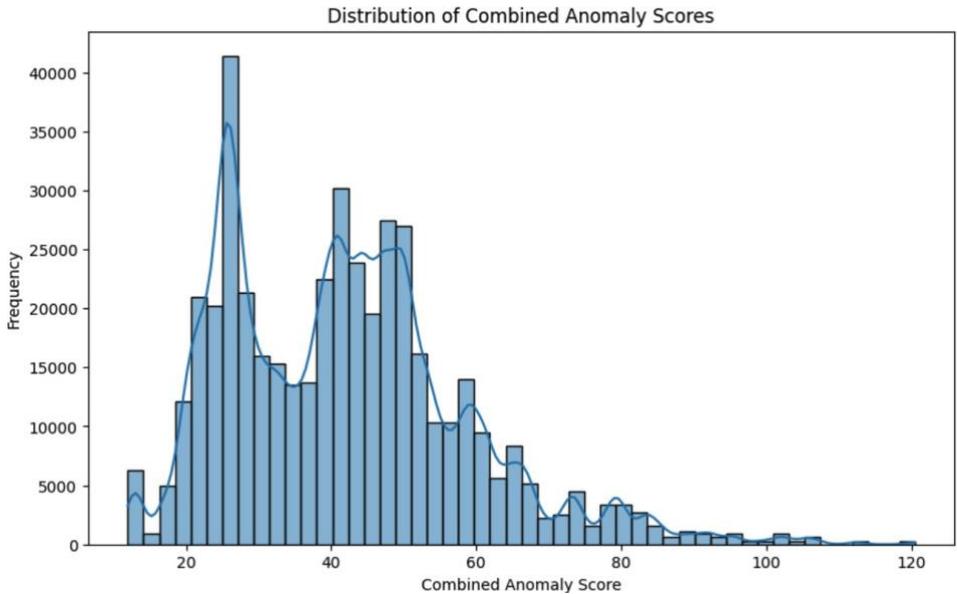

**Fig. 5**: Distribution of combined anomaly scores

### 3.5 Diversity and coverage analysis

The diversity and coverage analysis provided additional assurance regarding the dataset's ability to represent real-world variability. Demographic attributes, such as age, gender, and ethnicity, demonstrated a balanced distribution, ensuring that the synthetic dataset captured the diversity necessary for training robust AI models. Clinical variability was assessed through the representation of diagnoses, treatments, and medications, where the results indicated broad coverage across a wide range of clinical scenarios. This outcome was supported by high diversity indices, such as the Shannon Diversity Index, which confirmed the heterogeneity of the generated data. Additionally, category coverage analysis ensured that no significant underrepresented groups or fields were identified, thereby reinforcing the dataset's ability to generalize effectively across different patient populations and clinical contexts.

# 4 Discussion

The results of this study highlight the effectiveness of our framework in generating and validating synthetic healthcare datasets using generative AI models and advanced validation techniques. The foundation of this approach was prompt engineering with the GPT-4 API, which successfully guided the generation of realistic and structured synthetic data. The integration of this data into a relational PostgreSQL database, combined with rigorous validation, ensured the dataset achieved coherence, plausibility, consistency, and diversity which are key factors for training reliable AI models in healthcare. Numerous studies such as [22] and [23] have talked about the use of synthetic data in healthcare, but they do not demonstrate the comprehensive assess- ment of the quality of the synthetic data. In contrast, this study stands out because of the rigorous quality assessment of the generated data. Moreover, the large number of records (445,500) covering diverse clinical scenarios in our evaluations guarantees that the results of the analysis are reliable.

The Next Sentence Prediction (NSP) analysis demonstrated exceptional coherence across the synthetic records, with the majority of NSP probabilities concentrated at 1.00000 and the least recorded at 0.99800. These results, achieved through BERT's NSP model, affirm the logical progression and structure within clinical notes, medical histories, and other textual components, aligning with previous findings on the importance of sentence-level coherence in medical narratives [13]. This near-perfect coherence underscores the precision of prompt engineering in guiding GPT-4 to produce logically consistent records, which is critical in healthcare datasets where interdependent information must flow meaningfully.

The evaluation of plausibility using GPT-2 perplexity scores further validated the realism of the generated records. A total of 22,950 records exhibited perplexity scores below the predefined threshold, with most scores clustering between 20 and 50, reflecting strong adherence to natural medical text patterns. GPT-2's ability to assess language plausibility has been widely adopted [14], and the smooth distribu- tion observed in this study confirms that the generated records align with realistic linguistic conventions. While a small number of outliers exceeded the threshold, they highlight the challenges in synthesizing edge cases, consistent with prior studies that emphasize the difficulty of generating highly nuanced medical data [5].

The consistency analysis using RoBERTa's Natural Language Inference (NLI) model revealed strong logical alignment between related fields, with consistency scores predominantly ranging from 0.9875 to 0.9925 and a minimum score of 0.9750. These results reinforce the importance of domain-specific validation in healthcare, where relationships between symptoms, diagnoses, and treatments must follow established medical logic [15]. By identifying contradictions, such as mismatches between allergies and prescribed medications, the consistency checks ensured the dataset's integrity. High consistency across fields indicates that the synthetic records reflect medical rules and practices, making them suitable for downstream AI tasks.

The autoencoder-based anomaly detection identified 40,490 records with recon- struction errors exceeding the threshold. Despite these flagged anomalies, most errors were tightly clustered below 0.08, with only a small subset reaching 0.12, consistent with previous findings on anomaly detection in healthcare data [16]. Autoencoders effectively captured subtle deviations within numerical and categorical features, such as vital signs or lab results, which is critical for ensuring the reliability of high-dimensional synthetic datasets. The flagged records represent deviations that require iterative refinement but do not undermine the overall quality of the dataset.

The combined anomaly scores offered a holistic assessment of the dataset by integrating coherence, plausibility, and anomaly detection results. The scores ranged from 10 to 120, with the majority clustering between 20 and 60. This range indicates that most records exhibited minor to moderate deviations, while peaks observed between 30 and 40 reflect localized anomalies. This combined scoring approach aligns with studies demonstrating the value of integrating multiple validation metrics for synthetic data quality assurance [8].

The results of the diversity and coverage analysis confirmed the dataset's ability to reflect real-world variability. Demographic diversity was evident across attributes such as age, gender, and ethnicity, ensuring a balanced representation of patient populations. Clinical variability, measured through diagnoses, treatments, and medications, exhibited broad coverage, which was quantitatively supported by high Shannon Diversity Index scores. Ensuring diversity in synthetic datasets is critical for reducing biases and enhancing the generalizability of AI models [6]. By capturing this heterogeneity, the generated dataset avoids overfitting and improves the robustness of AI models applied to healthcare settings.

# 5 Conclusion

In conclusion, the results from NSP probabilities, perplexity scores, consistency analysis, anomaly detection, and diversity metrics demonstrate the effectiveness of our approach in generating high-quality synthetic healthcare data. The rigorous validation framework ensured coherence, plausibility, and logical consistency while maintaining sufficient diversity to reflect real-world variability. These findings align with existing literature on the importance of validation in synthetic data generation. The success- ful integration of validated data into the PostgreSQL database highlights its practical utility for Electronic Health Record (EHR) systems. By leveraging generative AI with advanced validation, this study offers a scalable, privacy-compliant solution for pro- ducing realistic synthetic healthcare data, addressing critical barriers in AI model training and development in healthcare.